\documentclass[10pt,twocolumn]{article} 
\usepackage{simpleConference}
\usepackage{times}
\usepackage{graphicx}
\usepackage{amssymb}
\PassOptionsToPackage{hyphens}{url}
\usepackage[hyphens]{url}
\usepackage{hyperref}
\hypersetup{
   breaklinks=true
}
\usepackage[numbers]{natbib}
\usepackage{subcaption}
\usepackage{amsthm}
\usepackage{amsbsy}
\usepackage{amsmath}
\usepackage{bbold}
\usepackage{multirow}
\usepackage{booktabs}
\usepackage{authblk}
\usepackage{color}
\usepackage{float}
\usepackage{pifont}
\usepackage{todonotes}
\usepackage{float}
\usepackage[capitalize]{cleveref}

\newsavebox{\tempfig}

\newcommand*{\plus}{%
    \raisebox{-0.01\baselineskip}{%
        \includegraphics[
        height=0.7\baselineskip,
        width=0.7\baselineskip,
        keepaspectratio,
        ]{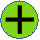}%
    }%
}
\newcommand*{\minus}{%
    \raisebox{-0.01\baselineskip}{%
        \includegraphics[
        height=0.7\baselineskip,
        width=0.7\baselineskip,
        keepaspectratio,
        ]{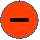}%
    }%
}

\title{Balancing Transparency and Risk: \\ The Security and Privacy Risks of Open-Source Machine Learning Models}

\makeatletter
\newcommand{\printfnsymbol}[1]{%
  \textsuperscript{\@fnsymbol{#1}}%
}
\makeatother

\author[1]{Dominik Hintersdorf\,\thanks{equal contribution}}
\author[1]{Lukas Struppek\,\printfnsymbol{1}}
\author[1,2,3,4]{Kristian Kersting}
\affil[1]{Technical University of Darmstadt, Germany}
\affil[2]{Hessian Center for AI (hessian.AI), Darmstadt, Germany}
\affil[3]{National Research Center for Applied Cybersecurity ATHENE, Darmstadt, Germany}
\affil[4]{German Research Center for Artificial Intelligence (DFKI), Darmstadt, Germany}
\affil[ ]{\textit{lastname@cs.tu-darmstadt.de}}

\begin{document}

\maketitle
\thispagestyle{empty}

\begin{abstract}
The field of artificial intelligence (AI) has experienced remarkable progress in recent years, driven by the widespread adoption of open-source machine learning models in both research and industry. Considering the resource-intensive nature of training on vast datasets, many applications opt for models that have already been trained. Hence, a small number of key players undertake the responsibility of training and publicly releasing large pre-trained models, providing a crucial foundation for a wide range of applications. However, the adoption of these open-source models carries inherent privacy and security risks that are often overlooked. To provide a concrete example, an inconspicuous model may conceal hidden functionalities that, when triggered by specific input patterns, can manipulate the behavior of the system, such as instructing self-driving cars to ignore the presence of other vehicles. The implications of successful privacy and security attacks encompass a broad spectrum, ranging from relatively minor damage like service interruptions to highly alarming scenarios, including physical harm or the exposure of sensitive user data. In this work, we present a comprehensive overview of common privacy and security threats associated with the use of open-source models. By raising awareness of these dangers, we strive to promote the responsible and secure use of AI systems. 

\end{abstract}

\section{Introduction}

\begin{figure}[t]
    \centering
    \includegraphics[width=\linewidth]{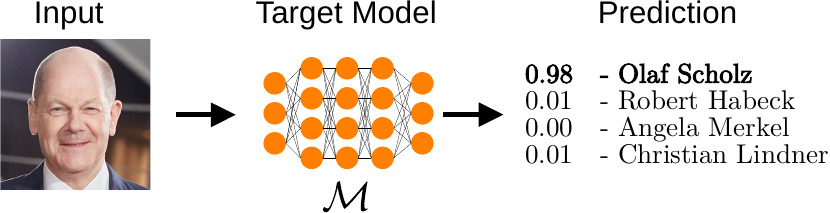}
    \caption{A basic deep neural network designed for facial recognition, capable of predicting corresponding identities, e.g., the German Chancellor Olaf Scholz. Given a specific input, the model computes a prediction vector, assigning probabilities to each distinct class. The final prediction is determined by the class with the highest probability. This model serves as an example for the attacks we discuss.}
    \label{fig:inferring_concept}
\end{figure}

With the increase of compute capability, big models are trained on a huge amount of data, often scraped from the public internet. Open-source models are often used as a basis for downstream tasks. As an example, the popular text-to-image model \textit{Stable Diffusion} uses the pre-trained text encoder from CLIP~\citep{dblp:clip}, a pre-trained multi-modal model, to process input texts.

While some large-scale models are completely closed-source, such as OpenAI's GPT-3~\cite{dblp:gpt3} or Google's Bard~\cite{dblp:lambda}, and are only accessible through an API, many other models are available as open-source models, usually including the code to train the model and the parameters of already trained models. Examples of such open-source models are BLOOM~\cite{dblp:bloom}, OpenLLaMA~\cite{openlm2023openllama}, LLaMA~\cite{dblp:llama}, LLaMA~2~\cite{arxiv:llama2}, OpenCLIP~\cite{open_clip} and Stable Diffusion~\cite{dblp:stable_diffusion}, and a group of companies, including GitHub, Hugging Face, Creative Commons, and others, are calling for more open-source support in the Forthcoming EU AI Act~\citep{verge23call}. While most open-source available models are trained on public data from the internet, information about which exact data was used is not always made public. Still, these models are deployed in numerous applications and settings.

But not only these big models are made publicly available. Sites like Hugging Face\footnote{\url{https://huggingface.co/}}, TensorFlow Hub\footnote{\url{https://tfhub.dev/}}, or PyTorch Hub\footnote{\url{https://pytorch.org/hub/}} allow users to provide and exchange model weights trained by the community, made publicly available to be downloaded by everyone. While this practice has clearly its upsides, the trustworthiness of such pre-trained open-source models comes increasingly into focus. Since the model architecture, weights and the training procedure are publicly known, malicious adversaries have an advantage when trying to attack these models compared to settings with models kept behind closed doors. Whereas all attacks presented in this work are also possible to some extent without full model access and less knowledge about the specific architecture, they become inherently more difficult to perform without such information.

Trustworthy machine learning comprises various areas, including security, safety, and privacy. Safety describes the robustness against model malfunctions without malicious external influences. For example, a \textit{safe} autonomous car provides reliable driving and transports people unharmed independent of the environmental conditions like weather. Security, on the other hand, describes a model's robustness against intentional attacks from malicious parties. For instance, an attacker could modify street signs to trigger a critical system behavior of the car and force a car crash. The aspect of \textit{privacy} relates to the access to private information about the models and their training data. Privacy-preserving models should not disclose any sensitive information from the training process to other users and attackers.

In this work, we will give an overview of common privacy and security threats associated with using open-source models.
In \cref{sec:privacy} and \cref{sec:security}, we will go over prominent privacy and security attacks. Then we will discuss the advantages and disadvantages of open-source practices in machine learning in \cref{sec:discussion}, followed by a conclusion in \cref{sec:conclusion}.

\section{Privacy Attacks on Open-Source Models}
\label{sec:privacy}
In this section, we will go over two most common privacy attacks, \textit{model inversion attacks} (cf. \cref{sec:model_inversion} and \cref{sec:extraction}) and \textit{membership inference attacks} (cf. \cref{sec:mem_inf}), and demonstrate how publicly releasing the model weights might harm user privacy. However, these attacks can also act as a tool to prevent unauthorized data usage. In the following, we will discuss both of these aspects of privacy attacks with regard to open-source models. \\

\subsection{Model Inversion Attacks}\label{sec:model_inversion}
\begin{figure}[t]
    \centering
    \includegraphics[width=\linewidth]{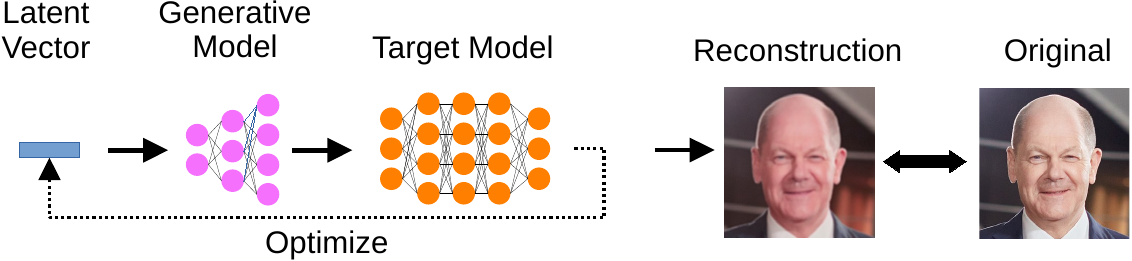}
    \caption{Model inversion attacks aim to synthesize samples that reveal sensitive information about the training data, such as revealing the identity of a person, in this case Olaf Scholz. The adversary usually employs a generative model, capable of producing synthetic images from a latent input vector. This latent vector is then optimized using the target model as guidance, with the objective of maximizing the confidence for a specific class.}
    \label{fig:model_inversion}
\end{figure}

Model inversion and reconstruction attacks have the goal of extracting sensitive information about the training data of an already trained model, e.g., by reconstructing images disclosing sensitive attributes~\cite{dblp:plug_and_play, dblp:vmi, dblp:ked, dblp:secret_revealer, dblp:fredrikson_mi, google_scholar:caia} or generating text with private information contained in the  training data~\cite{dblp:extracting_llm, dblp:canary_extraction}. \cref{fig:model_inversion} provides a simple example of a successful inversion attack.

For model inversion attack, it is usually assumed that the attacker has full access to the model and its parameters and also some generative model to generate samples from the training data domain. Generative models, in this case usually GANs~\citep{google_scholar:goodfellow_gans, dblp:stylegan}, are able to synthesize high-quality images from randomly sampled vectors, the so-called latent vectors. The generative model then acts as a prior to guide the optimization process and to generate images containing the sensitive features from the training data. Usually, the output value of a specific class of the model is maximized through an optimization process in which the latent vector of the generative model is altered. Even though, model inversion attacks are often applied to classification models, by altering the loss function of the optimization process these attacks can also be applied to models of other use cases.
As an attacker has full access to the open-source models, model inversion attacks are a genuine threat to the privacy of the training data. Imagine an open-source model trained to classify facial features like hair or eye color. An adversary successfully performing a model inversion attack could then generate synthetic facial images that reveal the identity of individuals from the training data.

\subsection{Information Leakage by Memorization}\label{sec:extraction}
Closely related to model inversion attacks is the issue of data leakage through unintended memorization. The distinction lies in the adversary's intent: in a model inversion attack, the adversary actively seeks to reconstruct model inputs, whereas leakage by memorization can occur incidentally, especially when interacting with generative models. These generative models encompass vast language models like the LLaMA family~\cite{dblp:llama, arxiv:llama2}, along with image generation models like Stable Diffusion~\cite{dblp:stable_diffusion}.

Generative language models, for instance, predict subsequent words when given a text input. For example, with the input sentence "the capital city of France is," a model might confidently predict "Paris." However, unintended leakage can happen when the model generates text containing private information from its training data that should not be disclosed as part of its prediction. For instance, the model might inadvertently complete the query "My social security number" with a real social security number that was present in the model's training data.

Since recent language models are trained on vast amounts of data scraped from various sources across the internet, it is highly probable that some private information will inadvertently become part of the model's training data. This highlights the importance of addressing and mitigating the risk of unintended data leakage, especially when dealing with generative models that have access to potentially sensitive information. In addition to accidental occurrences of memory leakage, there is also a concern that malicious users could deliberately craft queries that facilitate this kind of leakage~\citep{carlini19secret,lukas23analyzing}. This risk applies not only to open-source generative language models like LLaMA, but also to non-public models that offer only API access, potentially compromising individuals' privacy by generating texts containing sensitive information.

Likewise, similar concerns extend to image synthesis models, which have been found to reconstruct samples from their training data~\citep{vanderburg21memorization,carlini23diffextraction,somepalli23copying}. Such capabilities could potentially lead to legal issues if the generated content is under copyright protection. To address these challenges, it is crucial to implement robust privacy measures and security mechanisms in both language and image synthesis models, safeguarding against unintended data leakage and potential misuse of generated content. Proactive steps should be taken to mitigate the risks posed by both accidental and malicious attempts to exploit model vulnerabilities.

\subsection{Membership Inference Attacks}\label{sec:mem_inf}
\begin{figure}[t]
    \centering
    \includegraphics[width=\linewidth]{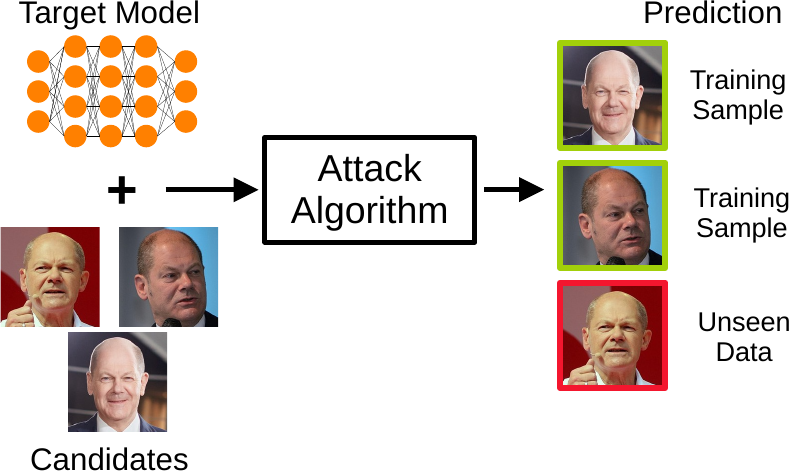}
    \caption{Membership Inference Attacks seek to determine whether a specific sample was part of a model's training data. These attacks commonly exploit that models tend to behave differently on inputs they have been trained on compared to unseen inputs.}
    \label{fig:mem_inf}
\end{figure}

While inversion and data leakage attacks try to infer information about the training data by reconstructing parts of it, membership inference attacks~\cite{dblp:meminf_shokri, dblp:meminf_to_trust, dblp:meminf_label_only_exposure, dblp:meminf_label_only_choquette, dblp:meminf_first_principles, dblp:meminf_yeom, dblp:meminf_salem}, as another type of privacy attack, try to infer which data samples have been used for training a model. \cref{fig:mem_inf} illustrates a simple example. In this scenario, the attacker has some data samples and wants to check whether this data was used for training a particular model. We will give a short example, to see why such a successful attack is a serious threat to privacy. Imagine that a hospital is training a machine learning model on the medical data of the hospital patients, to predict whether future patients have cancer. An attacker gains access to the model and has a set of private data samples. The adversary tries to infer whether the data of a person was used for training the cancer prediction model. If the attack is successful, the attacker knows not only that the person had or has cancer, but also was once a patient in that hospital. 
In the traditional setting of membership inference attacks, the attacker is interested in predicting whether a specific sample was present in the training data, i.e., a particular image or text. Related recent work such as from \citet{google_scholar:does_clip_know} or \citet{dblp:user_level_meminf} tries to infer if some data of a person was used for training without focussing on a particular data sample.

Having full access to an open-source model makes membership inference attacks more feasible in comparison to models kept behind APIs. This is because the attacker can observe the intermediate activations of every input, making it easier to infer membership.  As a result, open-source models can leak sensitive information about the data used for training. More importantly, this information about the training data is permanently encoded in the model weights. If private information is deleted from public websites, it is usually not publicly accessible anymore. However, if the model has been trained on this data, it still contains information about the data and can leak it to malicious users.

\subsection{Privacy Attacks to Enforce Rights}
Until now, we have only presented possible negative impacts of privacy attacks. However, there is also a positive side to open-source models being susceptible to these attacks. While these privacy attacks can leak possibly sensitive information to an attacker, they can also be used to prove unauthorized use of data. As a result, these attacks can be used to enforce privacy and copyright laws~\cite{google_scholar:does_clip_know}. 
Take for example the lawsuit of the stock image supplier Getty Images against Stability AI over copyright infringement. Getty Images accuses Stability AI of unlawfully using stock images for training their text-to-image model without having acquired a license to use the images~\cite{getty_images_lawsuit_statement, getty_images_lawsuit_verge}. Privacy attacks like model inversion, membership inference or memorization leakage attacks could be one way to prove that these images were illegally used for training.
Another example is that users can use these privacy attacks to prove that a company has trained a model on their potentially private data without permission, as shown by \citet{google_scholar:does_clip_know}. Combined with techniques to delete specific knowledge from the models~\cite{dblp:erasing_concepts_from_diffusion_models, dblp:forget_me_not} or machine unlearning~\cite{dblp:machine_unlearning}, these attacks offer a way to enforce the protection of user privacy.

\section{Security Attacks on Open-Source Models}\label{sec:security}
In this section, we show common security attacks against machine learning models. We will showcase two of the most prominent attack types, namely \textit{backdoor attacks} (cf. \cref{sec:backdoor_attacks} and \textit{adversarial examples} (cf. \cref{sec:adv_examples}).

\subsection{Data Poisoning and Backdoor Attacks}\label{sec:backdoor_attacks}

\begin{figure}[t]
    \centering
    \includegraphics[width=0.9\linewidth]{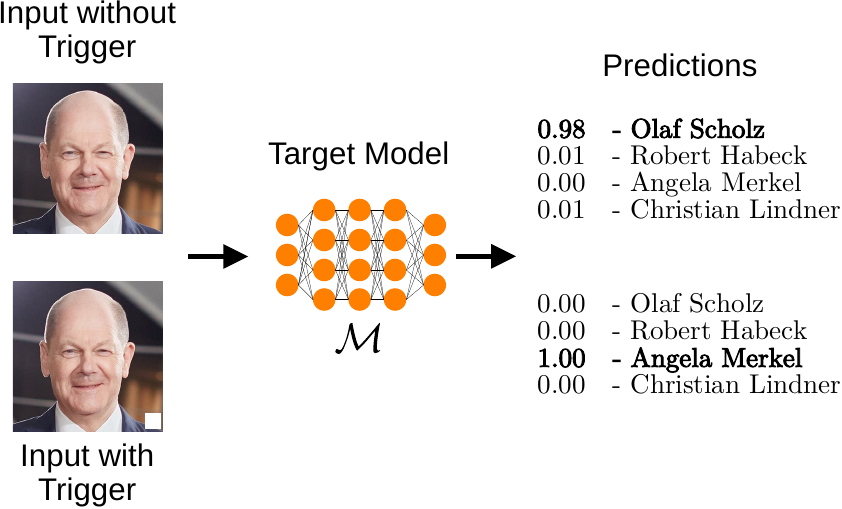}
    \caption{Backdoor attacks involve injecting a limited number of poisoned samples into a model's training data, aiming to inject a hidden model functionality, such as always predicting a specific class. During inference, this hidden behavior can be activated by inputs containing a pre-defined trigger, as illustrated in this example by a white square.}
    \label{fig:backdoor}
\end{figure}

Open-source models undergo training on vast datasets, often comprising millions or even billions of data samples. Due to this massive scale, human data inspection is not feasible in any way, necessitating a reliance on the integrity of these datasets. However, previous research has revealed that adding a small set of manipulated data to a model's training data can significantly influence its behavior. This dataset manipulation is referred to as \textit{data poisoning} and for numerous applications, manipulating less than 10\% of the available data is sufficient to make the model learn some additional, hidden functionalities.

Such hidden functionalities are called backdoors~\citep{gu2017, saha2020backdoor} and they are activated when the model input during inference includes a specific trigger pattern. \cref{fig:backdoor} demonstrates a practical backdoor attack. For instance, in the case of image classification, trigger patterns may involve certain color patterns placed in the corner of an image, e.g., a checkerboard pattern. A common backdoor strategy involves adding a small set of samples into the training data that contains both the trigger pattern and a target label from a particular class. During training, the model learns to associate the trigger pattern with the specified target class, thereby predicting the target class for each input that contains the trigger pattern. At the same time, the model's performance on clean inputs should not degrade noticeably to ensure the attack's stealthiness. 

Detecting this type of model manipulation is challenging for users since the models appear to function as expected on clean inputs. However, when the hidden backdoor function is activated, the model behaves as the attacker intended. A notable example are the text-to-image synthesis models, renowned for their ability to generate high-quality images based on textual descriptions provided by users. Nevertheless, Struppek {\it et al.}~\citep{struppek22rickrolling} have shown that small manipulations to the model are sufficient to inject backdoors that can be triggered by single characters or words. Once activated, these backdoors might force the generation of harmful or offensive content, posing serious risks to users. Depending on an individual's background, exposure to such generated content could cause mental harm and distress.

Backdoor and poisoning attacks have become prevalent across various machine learning domains, for example, image classification, self-supervised learning~\citep{carlini_backdoor_2022, Saha2022sslbackdoor}, transfer learning~\citep{yao2019latentbackdoor}, graph neural networks~\citep{xu2021expbackdoorgnn, zhang2021backdoor} and federated learning~\citep{zhang2022flbackdoor, shejwalkar2022flbackdoor}. There already exist various approaches to detect poisoned samples in the training data or triggers in the inputs. However, it is unclear if the training data of open-source models has been checked for poisoned data samples with existing approaches. Even if such inspections were conducted, providing an absolute guarantee that publicly available models are devoid of hidden backdoors remains challenging. The complexity and diversity of these attacks make it difficult to ensure complete protection.

\subsection{Adversarial Examples}\label{sec:adv_examples}
\begin{figure}[t]
    \centering
    \includegraphics[width=\linewidth]{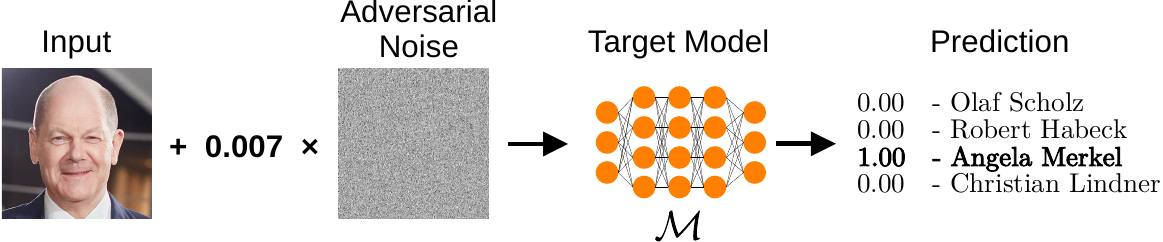}
    \caption{Adversarial examples are crafted by adding a small amount of fine-tuned noise to the input, which results in misleading predictions by the model. This adversarial noise is computed to alter the trained model's prediction in a specific manner. In many cases, the changes to the input are barely perceptible to humans, making it challenging to detect these manipulations.}
    \label{fig:adv_example}
\end{figure}

In addition to poisoning attacks that manipulate the training process to introduce hidden backdoor functions into a model, another category of security attacks targets models solely during inference. Known as adversarial examples or evasion attacks~\citep{goodfellow15adv}, these are slightly modified model inputs crafted with the intention of altering the model's behavior for the given input. Consequently, these adversarial examples can be employed to bypass a model's detection and cause misclassification of samples. \cref{fig:adv_example} illustrates a simple adversarial example. Among various security research subjects, adversarial examples stand out as the most extensively studied class of attacks, with numerous papers, amounting to several thousand, delving into this topic.

In computer vision tasks, the attacker computes a unique noise pattern tailored to a specific input, which is then added to the image to disrupt the model's prediction. Remarkably, even minor changes in the input, hardly noticeable to the human eye, can drastically impact the model's behavior. Numerous discussions have arisen concerning why deep learning architectures and other algorithms are susceptible to such subtle input changes. One plausible explanation lies in the models' dependence on non-robust input features that might not appear informative from a human standpoint. However, during training, these features can be exploited to solve the specific training task effectively~\cite{dblp:not_bugs_they_are_features}. 

In practice, adversarial examples are hard to detect by the human eye, rendering them especially dangerous in safety-critical applications. For instance, automatic content detection systems may be susceptible to evasion by images containing adversarial perturbations. This vulnerability extends to critical applications such as detecting child sexual abuse material~\citep{struppek_neuralhash} or identifying deepfakes~\citep{Hussain_2021_WACV}. The potential consequences of such undetected adversarial inputs emphasize the need to develop robust defenses against these attacks to ensure the integrity and reliability of machine learning systems.

Numerous approaches~\citep{goodfellow15adv, madry18pgd} to crafting adversarial examples leverage white-box model access, allowing them to compute gradients on the model concerning the current input. This enables the attacker to optimize the adversarial noise using standard gradient descent approaches. However, even with restricted access to a model's prediction vector~\citep{papernot17practical,pin17zoo} or only the predicted label~\citep{brendel18reliable,jianbo20hopskipjump}, various attack approaches still exist. The fact that open-source model weights and architectures are publicly available poses a risk, as adversaries can exploit the model locally and then use the crafted adversarial examples to deceive the targeted model. This highlights the importance of robust defense mechanisms to safeguard against such adversarial attacks, especially in scenarios when dealing with publicly accessible models.

\section{Discussion}
\label{sec:discussion}
While we have shown that publishing models as open-source has clearly disadvantages, there are also upsides to make models publicly available for everyone. In the following, we provide a discussion on both perspectives. Open-source machine learning models continue to be an important resource for the AI community despite these difficulties. By implementing best practices for model usage, performing security audits, and encouraging community cooperation to proactively solve security and privacy issues, risks can be reduced. Additionally, promoting responsible vulnerability disclosure can assist in preserving the security and dependability of open-source projects.

\noindent\minus~\textbf{Data Privacy Concerns:} Models trained on large datasets might inadvertently contain sensitive information, like personally identifiable information, medical data, or other sensitive details, posing privacy risks if not handled carefully. The models may inadvertently memorize or encode this information into its parameters during training. This can pose serious privacy risks when models are deployed in real-world applications. Samples from the training data could potentially be extracted through methods like model inversion attacks, allowing attackers to infer sensitive details about individuals whose data was used for training.

\noindent\minus~\textbf{Vulnerability Exposure:} Since open-source models are accessible to everyone, including malicious actors, vulnerabilities can be more easily exposed, potentially leading to strong attacks. Open-source models might become primary targets for adversarial attacks and evasion attacks. Malicious actors can study the model's architecture, parameters, and training data to develop sophisticated attacks aimed at manipulating or compromising the model's behavior.

\noindent\minus~\textbf{Lack of Regulatory Compliance \& License Issues:} Depending on the context of use, certain industries and applications might require compliance with specific security and privacy regulations. Using open-source models may complicate compliance efforts, especially if the model is not designed with these regulations in mind. Depending on the open-source license, some models may require users to disclose their modifications or share derived works, which could raise concerns about proprietary information. It is also an open question to which extent generative models can commit copyright infringement. Since parts of the training data may underlay copyright regulations, the generated data might also incorporate parts of it and, therefore, fall under copyright law.

\noindent\minus~\textbf{Zero-Day Vulnerabilities:} Open-source models can be susceptible to poisoning and backdoor attacks, where adversarial actors inject malicious data into the training set to manipulate the model's behavior. Many open-source models are published without their training data available. This makes it hard to check the integrity of the data and avoid model tampering of any kind. In practice, injected backdoors are hard to detect and may stay hidden until activated by a pre-defined trigger.

\noindent\plus~\textbf{Transparency and Auditability:} Open-source models allow users to examine the source code, algorithms, and sometimes even the data used to build the model. This transparency helps in understanding how the model works and detecting potential vulnerabilities. This process is called \textit{red-teaming} and is usually done by teams of the publishing companies such as OpenAI, Meta, or Google. In the case of open-source models, this process of finding and disclosing vulnerabilities can be done by the community in a much more open and transparent way.

\noindent\plus~\textbf{Community and Research Collaborations:} Open-source models encourage collaboration among researchers and developers. The community can work together to identify and fix security and privacy issues promptly. Furthermore, with access to novel models and architectures, existing attack and defense mechanisms can be investigated in this setting and allow adaptation and adjustments to new situations.

\noindent\plus~\textbf{Customization and Adaptation:} With access to the source code, developers can customize and adapt the model to suit their specific needs, ensuring it aligns with their security and privacy requirements. Since the available models are already trained, fewer data is required to adjust a model to a novel task or setting. In turn, fewer privacy concerns are expected from the fine-tuning dataset.

\noindent\plus~\textbf{Quality and Peer Review:} Popular open-source models often go through rigorous peer review, enhancing their overall quality and reducing the chances of major security or privacy flaws. It also includes investigations of independent research groups, offering new perspectives and insights.

\noindent\plus~\textbf{Faster Development and Innovation:} Building on top of existing open-source models can significantly speed up development efforts, enabling rapid innovation and research. This also includes the investigation of potential security vulnerabilities and corresponding defense and mitigation mechanisms.

\section{Conclusion}
\label{sec:conclusion}
In conclusion, we have highlighted and discussed the vulnerabilities of open-source models concerning security and privacy attacks, which are expected to pose a greater risk compared to closed-source models. The public access to model weights can significantly facilitate privacy attacks like inversion or membership inference, particularly when the training set remains private. Similarly, security attacks aimed at compromising model robustness can be executed by manipulating the training data to introduce hidden backdoor functionalities or crafting adversarial examples to manipulate inference outcomes. These risks not only impact the published model itself, but also extend to applications and systems that incorporate this model.

Despite these identified risks, it is important to acknowledge the numerous advantages that open-source machine learning offers. The practice of publishing models, source code, and potentially even data can support widespread adoption, foster transparency, and encourage innovation. We recognize the need for users and publishers to be aware of the inherent risks associated with open-source practices. However, particularly in the case of publishing large models, such as large language and text-to-image synthesis models, we firmly believe that the benefits outweigh the drawbacks. As such, we encourage developers to continue embracing open-source approaches, thereby promoting transparency, driving further research, and fostering innovation in the field of machine learning.

\medskip
{\bf Acknowledgments.}
The authors thank Daniel Neider for the fruitful discussions. This work was supported by the German Ministry of Education and Research (BMBF) within the framework program ``Research for Civil Security'' of the German Federal Government, project KISTRA (reference no. 13N15343).

\medskip

\textbf{Image sources:} Images depicting Olaf Scholz were provided by \citet{scholz_example}, \citet{scholz_example_2} and \citet{scholz_example_3}.

\newpage
\bibliographystyle{plainnat}
\bibliography{refs}

\begin{thebibliography}{61}
\providecommand{\natexlab}[1]{#1}
\providecommand{\url}[1]{\texttt{#1}}
\expandafter\ifx\csname urlstyle\endcsname\relax
  \providecommand{\doi}[1]{doi: #1}\else
  \providecommand{\doi}{doi: \begingroup \urlstyle{rm}\Url}\fi

\bibitem[Bourtoule et~al.(2021)Bourtoule, Chandrasekaran, Choquette{-}Choo,
  Jia, Travers, Zhang, Lie, and Papernot]{dblp:machine_unlearning}
Lucas Bourtoule, Varun Chandrasekaran, Christopher~A. Choquette{-}Choo, Hengrui
  Jia, Adelin Travers, Baiwu Zhang, David Lie, and Nicolas Papernot.
\newblock Machine unlearning.
\newblock In \emph{Symposium on Security and Privacy (S\&P)}, pages 141--159,
  2021.

\bibitem[Brendel et~al.(2018)Brendel, Rauber, and Bethge]{brendel18reliable}
Wieland Brendel, Jonas Rauber, and Matthias Bethge.
\newblock Decision-based adversarial attacks: Reliable attacks against
  black-box machine learning models.
\newblock In \emph{International Conference on Learning Representations
  (ICLR)}, 2018.

\bibitem[Brown et~al.(2020)Brown, Mann, Ryder, Subbiah, Kaplan, Dhariwal,
  Neelakantan, Shyam, Sastry, Askell, Agarwal, Herbert{-}Voss, Krueger,
  Henighan, Child, Ramesh, Ziegler, Wu, Winter, Hesse, Chen, Sigler, Litwin,
  Gray, Chess, Clark, Berner, McCandlish, Radford, Sutskever, and
  Amodei]{dblp:gpt3}
Tom~B. Brown, Benjamin Mann, Nick Ryder, Melanie Subbiah, Jared Kaplan,
  Prafulla Dhariwal, Arvind Neelakantan, Pranav Shyam, Girish Sastry, Amanda
  Askell, Sandhini Agarwal, Ariel Herbert{-}Voss, Gretchen Krueger, Tom
  Henighan, Rewon Child, Aditya Ramesh, Daniel~M. Ziegler, Jeffrey Wu, Clemens
  Winter, Christopher Hesse, Mark Chen, Eric Sigler, Mateusz Litwin, Scott
  Gray, Benjamin Chess, Jack Clark, Christopher Berner, Sam McCandlish, Alec
  Radford, Ilya Sutskever, and Dario Amodei.
\newblock Language models are few-shot learners.
\newblock In \emph{Advances in Neural Information Processing Systems
  (NeurIPS)}, 2020.

\bibitem[Carlini and Terzis(2022)]{carlini_backdoor_2022}
Nicholas Carlini and Andreas Terzis.
\newblock Poisoning and backdooring contrastive learning.
\newblock In \emph{International Conference on Learning Representations
  ({ICLR})}, 2022.

\bibitem[Carlini et~al.(2019)Carlini, Liu, Erlingsson, Kos, and
  Song]{carlini19secret}
Nicholas Carlini, Chang Liu, {\'{U}}lfar Erlingsson, Jernej Kos, and Dawn Song.
\newblock The secret sharer: Evaluating and testing unintended memorization in
  neural networks.
\newblock In \emph{{USENIX} Security Symposium}, pages 267--284, 2019.

\bibitem[Carlini et~al.(2021)Carlini, Tram{\`{e}}r, Wallace, Jagielski,
  Herbert{-}Voss, Lee, Roberts, Brown, Song, Erlingsson, Oprea, and
  Raffel]{dblp:extracting_llm}
Nicholas Carlini, Florian Tram{\`{e}}r, Eric Wallace, Matthew Jagielski, Ariel
  Herbert{-}Voss, Katherine Lee, Adam Roberts, Tom~B. Brown, Dawn Song,
  {\'{U}}lfar Erlingsson, Alina Oprea, and Colin Raffel.
\newblock Extracting training data from large language models.
\newblock In \emph{{USENIX} Security Symposium}, pages 2633--2650, 2021.

\bibitem[Carlini et~al.(2022)Carlini, Chien, Nasr, Song, Terzis, and
  Tram{\`{e}}r]{dblp:meminf_first_principles}
Nicholas Carlini, Steve Chien, Milad Nasr, Shuang Song, Andreas Terzis, and
  Florian Tram{\`{e}}r.
\newblock Membership inference attacks from first principles.
\newblock In \emph{Symposium on Security and Privacy (S\&P)}, pages 1897--1914,
  2022.

\bibitem[Carlini et~al.(2023)Carlini, Hayes, Nasr, Jagielski, Sehwag,
  Tram{\`{e}}r, Balle, Ippolito, and Wallace]{carlini23diffextraction}
Nicholas Carlini, Jamie Hayes, Milad Nasr, Matthew Jagielski, Vikash Sehwag,
  Florian Tram{\`{e}}r, Borja Balle, Daphne Ippolito, and Eric Wallace.
\newblock Extracting training data from diffusion models.
\newblock \emph{arXiv preprint}, arXiv:2301.13188, 2023.

\bibitem[Chen et~al.(2020)Chen, Jordan, and Wainwright]{jianbo20hopskipjump}
Jianbo Chen, Michael~I. Jordan, and Martin~J. Wainwright.
\newblock Hopskipjumpattack: {A} query-efficient decision-based attack.
\newblock In \emph{Symposium on Security and Privacy (S\&P)}, pages 1277--1294,
  2020.

\bibitem[Chen et~al.(2017)Chen, Zhang, Sharma, Yi, and Hsieh]{pin17zoo}
Pin{-}Yu Chen, Huan Zhang, Yash Sharma, Jinfeng Yi, and Cho{-}Jui Hsieh.
\newblock {ZOO:} zeroth order optimization based black-box attacks to deep
  neural networks without training substitute models.
\newblock In \emph{{ACM} Workshop on Artificial Intelligence and Security
  (AISec@CCS)}, pages 15--26, 2017.

\bibitem[Chen et~al.(2021)Chen, Kahla, Jia, and Qi]{dblp:ked}
Si~Chen, Mostafa Kahla, Ruoxi Jia, and Guo{-}Jun Qi.
\newblock Knowledge-enriched distributional model inversion attacks.
\newblock In \emph{International Conference on Computer Vision (ICCV)}, pages
  16158--16167, 2021.

\bibitem[Choquette{-}Choo et~al.(2021)Choquette{-}Choo, Tram{\`{e}}r, Carlini,
  and Papernot]{dblp:meminf_label_only_choquette}
Christopher~A. Choquette{-}Choo, Florian Tram{\`{e}}r, Nicholas Carlini, and
  Nicolas Papernot.
\newblock Label-only membership inference attacks.
\newblock In \emph{International Conference on Machine Learning (ICML)}, pages
  1964--1974, 2021.

\bibitem[David(2023)]{verge23call}
Emilia David.
\newblock Github and others call for more open-source support in eu ai law,
  2023.
\newblock
  https://www.theverge.com/2023/7/26/23807218/github-ai-open-source-creative-commons-hugging-face-eu-regulations,
  accessed: 27.07.2023.

\bibitem[Fredrikson et~al.(2015)Fredrikson, Jha, and
  Ristenpart]{dblp:fredrikson_mi}
Matt Fredrikson, Somesh Jha, and Thomas Ristenpart.
\newblock Model inversion attacks that exploit confidence information and basic
  countermeasures.
\newblock In \emph{{SIGSAC} Conference on Computer and Communications
  Security}, pages 1322--1333, 2015.

\bibitem[Gandikota et~al.(2023)Gandikota, Materzynska, Fiotto{-}Kaufman, and
  Bau]{dblp:erasing_concepts_from_diffusion_models}
Rohit Gandikota, Joanna Materzynska, Jaden Fiotto{-}Kaufman, and David Bau.
\newblock Erasing concepts from diffusion models.
\newblock \emph{arXiv preprint}, arXiv:2303.07345, 2023.

\bibitem[Geng and Liu(2023)]{openlm2023openllama}
Xinyang Geng and Hao Liu.
\newblock Openllama: An open reproduction of llama, 2023.
\newblock URL \url{https://github.com/openlm-research/open_llama}.

\bibitem[Goodfellow et~al.(2014)Goodfellow, Pouget-Abadie, Mirza, Xu,
  Warde-Farley, Ozair, Courville, and Bengio]{google_scholar:goodfellow_gans}
Ian Goodfellow, Jean Pouget-Abadie, Mehdi Mirza, Bing Xu, David Warde-Farley,
  Sherjil Ozair, Aaron Courville, and Yoshua Bengio.
\newblock Generative adversarial nets.
\newblock In \emph{Advances in Neural Information Processing Systems
  (NeurIPS)}, 2014.

\bibitem[Goodfellow et~al.(2015)Goodfellow, Shlens, and
  Szegedy]{goodfellow15adv}
Ian~J. Goodfellow, Jonathon Shlens, and Christian Szegedy.
\newblock Explaining and harnessing adversarial examples.
\newblock In \emph{International Conference on Learning Representations
  ({ICLR})}, 2015.

\bibitem[Gu et~al.(2017)Gu, Dolan{-}Gavitt, and Garg]{gu2017}
Tianyu Gu, Brendan Dolan{-}Gavitt, and Siddharth Garg.
\newblock Badnets: Identifying vulnerabilities in the machine learning model
  supply chain.
\newblock \emph{arXiv preprint}, arXiv:1708.06733, 2017.

\bibitem[Heinrich-Böll-Stiftung()]{scholz_example_3}
Heinrich-Böll-Stiftung.
\newblock
  \url{https://commons.wikimedia.org/wiki/File:Olaf_Scholz_(14271189360).jpg},
  Licensed as CC BY-SA 2.0, accessed: 24.07.2023.

\bibitem[Hintersdorf et~al.(2022{\natexlab{a}})Hintersdorf, Struppek, and
  Kersting]{dblp:meminf_to_trust}
Dominik Hintersdorf, Lukas Struppek, and Kristian Kersting.
\newblock To trust or not to trust prediction scores for membership inference
  attacks.
\newblock In \emph{International Joint Conference on Artificial Intelligence
  (IJCAI)}, pages 3043--3049, 2022{\natexlab{a}}.

\bibitem[Hintersdorf et~al.(2022{\natexlab{b}})Hintersdorf, Struppek, and
  Kersting]{google_scholar:does_clip_know}
Dominik Hintersdorf, Lukas Struppek, and Kristian Kersting.
\newblock Clipping privacy: Identity inference attacks on multi-modal machine
  learning models.
\newblock \emph{arXiv preprint}, arXiv:2209.07341, 2022{\natexlab{b}}.

\bibitem[Hussain et~al.(2021)Hussain, Neekhara, Jere, Koushanfar, and
  McAuley]{Hussain_2021_WACV}
Shehzeen Hussain, Paarth Neekhara, Malhar Jere, Farinaz Koushanfar, and Julian
  McAuley.
\newblock Adversarial deepfakes: Evaluating vulnerability of deepfake detectors
  to adversarial examples.
\newblock In \emph{Winter Conference on Applications of Computer Vision
  (WACV)}, pages 3348--3357, 2021.

\bibitem[Ilharco et~al.(2021)Ilharco, Wortsman, Wightman, Gordon, Carlini,
  Taori, Dave, Shankar, Namkoong, Miller, Hajishirzi, Farhadi, and
  Schmidt]{open_clip}
Gabriel Ilharco, Mitchell Wortsman, Ross Wightman, Cade Gordon, Nicholas
  Carlini, Rohan Taori, Achal Dave, Vaishaal Shankar, Hongseok Namkoong, John
  Miller, Hannaneh Hajishirzi, Ali Farhadi, and Ludwig Schmidt.
\newblock Openclip, 2021.

\bibitem[Ilyas et~al.(2019)Ilyas, Santurkar, Tsipras, Engstrom, Tran, and
  Madry]{dblp:not_bugs_they_are_features}
Andrew Ilyas, Shibani Santurkar, Dimitris Tsipras, Logan Engstrom, Brandon
  Tran, and Aleksander Madry.
\newblock Adversarial examples are not bugs, they are features.
\newblock In \emph{Advances in Neural Information Processing Systems
  (NeurIPS)}, pages 125--136, 2019.

\bibitem[Images(2023)]{getty_images_lawsuit_statement}
Getty Images.
\newblock Getty images statement.
\newblock
  \url{https://newsroom.gettyimages.com/en/getty-images/getty-images-statement},
  2023.
\newblock Online; accessed 24-July-2023.

\bibitem[Karras et~al.(2019)Karras, Laine, and Aila]{dblp:stylegan}
Tero Karras, Samuli Laine, and Timo Aila.
\newblock A style-based generator architecture for generative adversarial
  networks.
\newblock In \emph{Conference on Computer Vision and Pattern Recognition
  (CVPR)}, pages 4401--4410, 2019.

\bibitem[Li et~al.(2022)Li, Rezaei, and Liu]{dblp:user_level_meminf}
Guoyao Li, Shahbaz Rezaei, and Xin Liu.
\newblock User-level membership inference attack against metric embedding
  learning.
\newblock \emph{arXiv preprint}, arXiv:2203.02077, 2022.

\bibitem[Li and Zhang(2021)]{dblp:meminf_label_only_exposure}
Zheng Li and Yang Zhang.
\newblock Membership leakage in label-only exposures.
\newblock In \emph{Conference on Computer and Communications Security (CCS)},
  pages 880--895, 2021.

\bibitem[Lucan()]{scholz_example_2}
Michael Lucan.
\newblock
  \url{https://commons.wikimedia.org/wiki/File:2021-08-21_Olaf_Scholz_0433.JPG},
  Licensed as CC BY-SA 3.0, accessed: 24.07.2023.

\bibitem[Ludewig()]{scholz_example}
Bernhard Ludewig.
\newblock \url{https://www.flickr.com/photos/finnishgovernment/51941396612/},
  Licensed as CC BY 2.0, accessed: 24.07.2023.

\bibitem[Lukas et~al.(2023)Lukas, Salem, Sim, Tople, Wutschitz, and
  Zanella-Béguelin]{lukas23analyzing}
Nils Lukas, Ahmed Salem, Robert Sim, Shruti Tople, Lukas Wutschitz, and
  Santiago Zanella-Béguelin.
\newblock Analyzing leakage of personally identifiable information in language
  models.
\newblock In \emph{Symposium on Security and Privacy (S\&P)}, pages 346--363,
  2023.

\bibitem[Madry et~al.(2018)Madry, Makelov, Schmidt, Tsipras, and
  Vladu]{madry18pgd}
Aleksander Madry, Aleksandar Makelov, Ludwig Schmidt, Dimitris Tsipras, and
  Adrian Vladu.
\newblock Towards deep learning models resistant to adversarial attacks.
\newblock In \emph{International Conference on Learning Representations
  ({ICLR})}, 2018.

\bibitem[Papernot et~al.(2017)Papernot, McDaniel, Goodfellow, Jha, Celik, and
  Swami]{papernot17practical}
Nicolas Papernot, Patrick McDaniel, Ian Goodfellow, Somesh Jha, Z.~Berkay
  Celik, and Ananthram Swami.
\newblock Practical black-box attacks against machine learning.
\newblock In \emph{Asia Conference on Computer and Communications Security
  (ASIA CCS)}, page 506–519, 2017.

\bibitem[Parikh et~al.(2022)Parikh, Dupuy, and Gupta]{dblp:canary_extraction}
Rahil Parikh, Christophe Dupuy, and Rahul Gupta.
\newblock Canary extraction in natural language understanding models.
\newblock In \emph{Annual Meeting of the Association for Computational
  Linguistics ({ACL}) - Short Paper}, pages 552--560, 2022.

\bibitem[Radford et~al.(2021)Radford, Kim, Hallacy, Ramesh, Goh, Agarwal,
  Sastry, Askell, Mishkin, Clark, Krueger, and Sutskever]{dblp:clip}
Alec Radford, Jong~Wook Kim, Chris Hallacy, Aditya Ramesh, Gabriel Goh,
  Sandhini Agarwal, Girish Sastry, Amanda Askell, Pamela Mishkin, Jack Clark,
  Gretchen Krueger, and Ilya Sutskever.
\newblock Learning transferable visual models from natural language
  supervision.
\newblock In \emph{International Conference on Machine Learning (ICML)}, pages
  8748--8763, 2021.

\bibitem[Rombach et~al.(2022)Rombach, Blattmann, Lorenz, Esser, and
  Ommer]{dblp:stable_diffusion}
Robin Rombach, Andreas Blattmann, Dominik Lorenz, Patrick Esser, and
  Bj{\"{o}}rn Ommer.
\newblock High-resolution image synthesis with latent diffusion models.
\newblock In \emph{Conference on Computer Vision and Pattern Recognition
  (CVPR)}, pages 10674--10685, 2022.

\bibitem[Saha et~al.(2020)Saha, Subramanya, and Pirsiavash]{saha2020backdoor}
Aniruddha Saha, Akshayvarun Subramanya, and Hamed Pirsiavash.
\newblock Hidden trigger backdoor attacks.
\newblock In \emph{Conference on Artificial Intelligence (AAAI)}, pages
  11957--11965, 2020.

\bibitem[Saha et~al.(2022)Saha, Tejankar, Koohpayegani, and
  Pirsiavash]{Saha2022sslbackdoor}
Aniruddha Saha, Ajinkya Tejankar, Soroush~Abbasi Koohpayegani, and Hamed
  Pirsiavash.
\newblock Backdoor attacks on self-supervised learning.
\newblock In \emph{Conference on Computer Vision and Pattern Recognition
  (CVPR)}, pages 13337--13346, 2022.

\bibitem[Salem et~al.(2019)Salem, Zhang, Humbert, Berrang, Fritz, and
  Backes]{dblp:meminf_salem}
Ahmed Salem, Yang Zhang, Mathias Humbert, Pascal Berrang, Mario Fritz, and
  Michael Backes.
\newblock Ml-leaks: Model and data independent membership inference attacks and
  defenses on machine learning models.
\newblock In \emph{Annual Network and Distributed System Security Symposium
  ({NDSS})}, 2019.

\bibitem[Scao et~al.(2022)Scao, Fan, Akiki, Pavlick, Ilic, Hesslow,
  Castagn{\'{e}}, Luccioni, Yvon, Gall{\'{e}}, Tow, Rush, Biderman, Webson,
  Ammanamanchi, Wang, Sagot, Muennighoff, del Moral, Ruwase, Bawden, Bekman,
  McMillan{-}Major, Beltagy, Nguyen, Saulnier, Tan, Suarez, Sanh,
  Lauren{\c{c}}on, Jernite, Launay, Mitchell, Raffel, Gokaslan, Simhi, Soroa,
  Aji, Alfassy, Rogers, Nitzav, Xu, Mou, Emezue, Klamm, Leong, van Strien,
  Adelani, and et~al.]{dblp:bloom}
Teven~Le Scao, Angela Fan, Christopher Akiki, Ellie Pavlick, Suzana Ilic,
  Daniel Hesslow, Roman Castagn{\'{e}}, Alexandra~Sasha Luccioni,
  Fran{\c{c}}ois Yvon, Matthias Gall{\'{e}}, Jonathan Tow, Alexander~M. Rush,
  Stella Biderman, Albert Webson, Pawan~Sasanka Ammanamanchi, Thomas Wang,
  Beno{\^{\i}}t Sagot, Niklas Muennighoff, Albert~Villanova del Moral, Olatunji
  Ruwase, Rachel Bawden, Stas Bekman, Angelina McMillan{-}Major, Iz~Beltagy,
  Huu Nguyen, Lucile Saulnier, Samson Tan, Pedro~Ortiz Suarez, Victor Sanh,
  Hugo Lauren{\c{c}}on, Yacine Jernite, Julien Launay, Margaret Mitchell, Colin
  Raffel, Aaron Gokaslan, Adi Simhi, Aitor Soroa, Alham~Fikri Aji, Amit
  Alfassy, Anna Rogers, Ariel~Kreisberg Nitzav, Canwen Xu, Chenghao Mou, Chris
  Emezue, Christopher Klamm, Colin Leong, Daniel van Strien, David~Ifeoluwa
  Adelani, and et~al.
\newblock {BLOOM:} {A} 176b-parameter open-access multilingual language model.
\newblock \emph{arXiv preprint}, arXiv:2211.05100, 2022.

\bibitem[Shejwalkar et~al.(2022)Shejwalkar, Houmansadr, Kairouz, and
  Ramage]{shejwalkar2022flbackdoor}
Virat Shejwalkar, Amir Houmansadr, Peter Kairouz, and Daniel Ramage.
\newblock Back to the drawing board: {A} critical evaluation of poisoning
  attacks on production federated learning.
\newblock In \emph{Symposium on Security and Privacy (S\&P)}, pages 1354--1371,
  2022.

\bibitem[Shokri et~al.(2017)Shokri, Stronati, Song, and
  Shmatikov]{dblp:meminf_shokri}
Reza Shokri, Marco Stronati, Congzheng Song, and Vitaly Shmatikov.
\newblock Membership inference attacks against machine learning models.
\newblock In \emph{Symposium on Security and Privacy (S\&P)}, pages 3--18,
  2017.

\bibitem[Somepalli et~al.(2023)Somepalli, Singla, Goldblum, Geiping, and
  Goldstein]{somepalli23copying}
Gowthami Somepalli, Vasu Singla, Micah Goldblum, Jonas Geiping, and Tom
  Goldstein.
\newblock Understanding and mitigating copying in diffusion models.
\newblock \emph{arXiv preprint}, arXiv:2305.20086, 2023.

\bibitem[Struppek et~al.(2022{\natexlab{a}})Struppek, Hintersdorf, Correia,
  Adler, and Kersting]{dblp:plug_and_play}
Lukas Struppek, Dominik Hintersdorf, Antonio De~Almeida Correia, Antonia Adler,
  and Kristian Kersting.
\newblock Plug {\&} play attacks: Towards robust and flexible model inversion
  attacks.
\newblock In \emph{International Conference on Machine Learning (ICML)}, pages
  20522--20545, 2022{\natexlab{a}}.

\bibitem[Struppek et~al.(2022{\natexlab{b}})Struppek, Hintersdorf, Neider, and
  Kersting]{struppek_neuralhash}
Lukas Struppek, Dominik Hintersdorf, Daniel Neider, and Kristian Kersting.
\newblock Learning to break deep perceptual hashing: The use case neuralhash.
\newblock In \emph{Conference on Fairness, Accountability, and Transparency
  (FAccT)}, page 58–69, 2022{\natexlab{b}}.

\bibitem[Struppek et~al.(2023{\natexlab{a}})Struppek, Hintersdorf, Friedrich,
  Brack, Schramowski, and Kersting]{google_scholar:caia}
Lukas Struppek, Dominik Hintersdorf, Felix Friedrich, Manuel Brack, Patrick
  Schramowski, and Kristian Kersting.
\newblock Image classifiers leak sensitive attributes about their classes.
\newblock \emph{arXiv preprint}, arXiv:2303.09289, 2023{\natexlab{a}}.

\bibitem[Struppek et~al.(2023{\natexlab{b}})Struppek, Hintersdorf, and
  Kersting]{struppek22rickrolling}
Lukas Struppek, Dominik Hintersdorf, and Kristian Kersting.
\newblock Rickrolling the artist: Injecting backdoors into text-guided image
  generation models.
\newblock In \emph{International Conference on Computer Vision (ICCV)},
  2023{\natexlab{b}}.

\bibitem[Thoppilan et~al.(2022)Thoppilan, Freitas, Hall, Shazeer, Kulshreshtha,
  Cheng, Jin, Bos, Baker, Du, Li, Lee, Zheng, Ghafouri, Menegali, Huang,
  Krikun, Lepikhin, Qin, Chen, Xu, Chen, Roberts, Bosma, Zhou, Chang, Krivokon,
  Rusch, Pickett, Meier{-}Hellstern, Morris, Doshi, Santos, Duke, Soraker,
  Zevenbergen, Prabhakaran, Diaz, Hutchinson, Olson, Molina, Hoffman{-}John,
  Lee, Aroyo, Rajakumar, Butryna, Lamm, Kuzmina, Fenton, Cohen, Bernstein,
  Kurzweil, y~Arcas, Cui, Croak, Chi, and Le]{dblp:lambda}
Romal Thoppilan, Daniel~De Freitas, Jamie Hall, Noam Shazeer, Apoorv
  Kulshreshtha, Heng{-}Tze Cheng, Alicia Jin, Taylor Bos, Leslie Baker, Yu~Du,
  YaGuang Li, Hongrae Lee, Huaixiu~Steven Zheng, Amin Ghafouri, Marcelo
  Menegali, Yanping Huang, Maxim Krikun, Dmitry Lepikhin, James Qin, Dehao
  Chen, Yuanzhong Xu, Zhifeng Chen, Adam Roberts, Maarten Bosma, Yanqi Zhou,
  Chung{-}Ching Chang, Igor Krivokon, Will Rusch, Marc Pickett, Kathleen~S.
  Meier{-}Hellstern, Meredith~Ringel Morris, Tulsee Doshi, Renelito~Delos
  Santos, Toju Duke, Johnny Soraker, Ben Zevenbergen, Vinodkumar Prabhakaran,
  Mark Diaz, Ben Hutchinson, Kristen Olson, Alejandra Molina, Erin
  Hoffman{-}John, Josh Lee, Lora Aroyo, Ravi Rajakumar, Alena Butryna, Matthew
  Lamm, Viktoriya Kuzmina, Joe Fenton, Aaron Cohen, Rachel Bernstein, Ray
  Kurzweil, Blaise~Ag{\"{u}}era y~Arcas, Claire Cui, Marian Croak, Ed~H. Chi,
  and Quoc Le.
\newblock Lamda: Language models for dialog applications.
\newblock \emph{arXiv preprint}, arXiv:2201.08239, 2022.

\bibitem[Touvron et~al.(2023{\natexlab{a}})Touvron, Lavril, Izacard, Martinet,
  Lachaux, Lacroix, Rozi{\`{e}}re, Goyal, Hambro, Azhar, Rodriguez, Joulin,
  Grave, and Lample]{dblp:llama}
Hugo Touvron, Thibaut Lavril, Gautier Izacard, Xavier Martinet, Marie{-}Anne
  Lachaux, Timoth{\'{e}}e Lacroix, Baptiste Rozi{\`{e}}re, Naman Goyal, Eric
  Hambro, Faisal Azhar, Aur{\'{e}}lien Rodriguez, Armand Joulin, Edouard Grave,
  and Guillaume Lample.
\newblock Llama: Open and efficient foundation language models.
\newblock \emph{arXiv preprint}, arXiv:2302.13971, 2023{\natexlab{a}}.

\bibitem[Touvron et~al.(2023{\natexlab{b}})Touvron, Martin, Stone, Albert,
  Almahairi, Babaei, Bashlykov, Batra, Bhargava, Bhosale, Bikel, Blecher,
  Ferrer, Chen, Cucurull, Esiobu, Fernandes, Fu, Fu, Fuller, Gao, Goswami,
  Goyal, Hartshorn, Hosseini, Hou, Inan, Kardas, Kerkez, Khabsa, Kloumann,
  Korenev, Koura, Lachaux, Lavril, Lee, Liskovich, Lu, Mao, Martinet, Mihaylov,
  Mishra, Molybog, Nie, Poulton, Reizenstein, Rungta, Saladi, Schelten, Silva,
  Smith, Subramanian, Tan, Tang, Taylor, Williams, Kuan, Xu, Yan, Zarov, Zhang,
  Fan, Kambadur, Narang, Rodriguez, Stojnic, Edunov, and Scialom]{arxiv:llama2}
Hugo Touvron, Louis Martin, Kevin Stone, Peter Albert, Amjad Almahairi, Yasmine
  Babaei, Nikolay Bashlykov, Soumya Batra, Prajjwal Bhargava, Shruti Bhosale,
  Dan Bikel, Lukas Blecher, Cristian~Canton Ferrer, Moya Chen, Guillem
  Cucurull, David Esiobu, Jude Fernandes, Jeremy Fu, Wenyin Fu, Brian Fuller,
  Cynthia Gao, Vedanuj Goswami, Naman Goyal, Anthony Hartshorn, Saghar
  Hosseini, Rui Hou, Hakan Inan, Marcin Kardas, Viktor Kerkez, Madian Khabsa,
  Isabel Kloumann, Artem Korenev, Punit~Singh Koura, Marie-Anne Lachaux,
  Thibaut Lavril, Jenya Lee, Diana Liskovich, Yinghai Lu, Yuning Mao, Xavier
  Martinet, Todor Mihaylov, Pushkar Mishra, Igor Molybog, Yixin Nie, Andrew
  Poulton, Jeremy Reizenstein, Rashi Rungta, Kalyan Saladi, Alan Schelten, Ruan
  Silva, Eric~Michael Smith, Ranjan Subramanian, Xiaoqing~Ellen Tan, Binh Tang,
  Ross Taylor, Adina Williams, Jian~Xiang Kuan, Puxin Xu, Zheng Yan, Iliyan
  Zarov, Yuchen Zhang, Angela Fan, Melanie Kambadur, Sharan Narang, Aurelien
  Rodriguez, Robert Stojnic, Sergey Edunov, and Thomas Scialom.
\newblock Llama 2: Open foundation and fine-tuned chat models.
\newblock \emph{arXiv preprint}, arXiv:2307.09288, 2023{\natexlab{b}}.

\bibitem[van~den Burg and Williams(2021)]{vanderburg21memorization}
Gerrit J.~J. van~den Burg and Chris Williams.
\newblock On memorization in probabilistic deep generative models.
\newblock In \emph{Advances in Neural Information Processing Systems
  (NeurIPS)}, pages 27916--27928, 2021.

\bibitem[Vincent(2023)]{getty_images_lawsuit_verge}
James Vincent.
\newblock Getty images is suing the creators of ai art tool stable diffusion
  for scraping its content.
\newblock
  \url{https://www.theverge.com/2023/1/17/23558516/ai-art-copyright-stable-diffusion-getty-images-lawsuit},
  2023.
\newblock Online; accessed 24-July-2023.

\bibitem[Wang et~al.(2021)Wang, Fu, Li, Khisti, Zemel, and Makhzani]{dblp:vmi}
Kuan{-}Chieh Wang, Yan Fu, Ke~Li, Ashish Khisti, Richard~S. Zemel, and Alireza
  Makhzani.
\newblock Variational model inversion attacks.
\newblock In \emph{Advances in Neural Information Processing Systems
  (NeurIPS)}, pages 9706--9719, 2021.

\bibitem[Xu et~al.(2021)Xu, Xue, and Picek]{xu2021expbackdoorgnn}
Jing Xu, Minhui Xue, and Stjepan Picek.
\newblock Explainability-based backdoor attacks against graph neural networks.
\newblock In \emph{{ACM} Workshop on Wireless Security and Machine Learning},
  pages 31--36, 2021.

\bibitem[Yao et~al.(2019)Yao, Li, Zheng, and Zhao]{yao2019latentbackdoor}
Yuanshun Yao, Huiying Li, Haitao Zheng, and Ben~Y. Zhao.
\newblock Latent backdoor attacks on deep neural networks.
\newblock In \emph{Conference on Computer and Communications Security (CCS)},
  pages 2041--2055, 2019.

\bibitem[Yeom et~al.(2018)Yeom, Giacomelli, Fredrikson, and
  Jha]{dblp:meminf_yeom}
Samuel Yeom, Irene Giacomelli, Matt Fredrikson, and Somesh Jha.
\newblock Privacy risk in machine learning: Analyzing the connection to
  overfitting.
\newblock In \emph{Computer Security Foundations Symposium (CSF)}, pages
  268--282, 2018.

\bibitem[Zhang et~al.(2023)Zhang, Wang, Xu, Wang, and Shi]{dblp:forget_me_not}
Eric Zhang, Kai Wang, Xingqian Xu, Zhangyang Wang, and Humphrey Shi.
\newblock Forget-me-not: Learning to forget in text-to-image diffusion models.
\newblock \emph{arXiv preprint}, arXiv:2303.17591, 2023.

\bibitem[Zhang et~al.(2020)Zhang, Jia, Pei, Wang, Li, and
  Song]{dblp:secret_revealer}
Yuheng Zhang, Ruoxi Jia, Hengzhi Pei, Wenxiao Wang, Bo~Li, and Dawn Song.
\newblock The secret revealer: Generative model-inversion attacks against deep
  neural networks.
\newblock In \emph{Conference on Computer Vision and Pattern Recognition
  (CVPR)}, pages 250--258, 2020.

\bibitem[Zhang et~al.(2021)Zhang, Jia, Wang, and Gong]{zhang2021backdoor}
Zaixi Zhang, Jinyuan Jia, Binghui Wang, and Neil~Zhenqiang Gong.
\newblock Backdoor attacks to graph neural networks.
\newblock In \emph{{ACM} Symposium on Access Control Models and Technologies
  (SACMAT)}, pages 15--26, 2021.

\bibitem[Zhang et~al.(2022)Zhang, Panda, Song, Yang, Mahoney, Mittal,
  Ramchandran, and Gonzalez]{zhang2022flbackdoor}
Zhengming Zhang, Ashwinee Panda, Linyue Song, Yaoqing Yang, Michael~W. Mahoney,
  Prateek Mittal, Kannan Ramchandran, and Joseph Gonzalez.
\newblock Neurotoxin: Durable backdoors in federated learning.
\newblock In \emph{International Conference on Machine Learning ({ICML})},
  volume 162, pages 26429--26446, 2022.

\end{thebibliography}

\end{document}